\pdfoutput=1
\documentclass[letterpaper]{article} 
\usepackage{aaai24}  
\usepackage{times}  
\usepackage{helvet}  
\usepackage{courier}  
\usepackage[hyphens]{url}  
\usepackage{graphicx} 
\usepackage{natbib}  
\usepackage{caption} 
\usepackage{algorithm}
\usepackage{algorithmic}

\usepackage{newfloat}
\usepackage{listings}
\DeclareCaptionStyle{ruled}{labelfont=normalfont,labelsep=colon,strut=off} 
\floatstyle{ruled}
\newfloat{listing}{tb}{lst}{}
\floatname{listing}{Listing}
\usepackage{bm}
\usepackage{ragged2e} 
\usepackage{booktabs,makecell, multirow, tabularx}
\usepackage{amsmath}
\usepackage{amssymb}

\title{IS-DARTS: Stabilizing DARTS through Precise Measurement\\ on Candidate Importance}
\author {
    Hongyi He,
    Longjun Liu\thanks{Longjun Liu is the corresponding author.},
    Haonan Zhang,
    Nanning Zheng
}
\affiliations {
    National Key Laboratory of Human-Machine Hybrid Augmented Intelligence, National Engineering Research Center for Visual Information and Applications, and Institute of Artificial Intelligence and Robotics, Xi'an Jiaotong University\\
    hongyihe@stu.xjtu.edu.cn, liulongjun@xjtu.edu.cn, haonanzhang@stu.xjtu.edu.cn, nnzheng@xjtu.edu.cn
}

\usepackage{bibentry}

\begin{document}

\maketitle

\begin{abstract}
Among existing Neural Architecture Search methods, DARTS is known for its efficiency and simplicity. This approach applies continuous relaxation of network representation to construct a weight-sharing supernet and enables the identification of excellent subnets in just a few GPU days. However, performance collapse in DARTS results in deteriorating architectures filled with parameter-free operations and remains a great challenge to the robustness. 
To resolve this problem, we reveal that the fundamental reason is the biased estimation of the candidate importance in the search space through theoretical and experimental analysis, and more precisely select operations via information-based measurements. 
Furthermore, we demonstrate that the excessive concern over the supernet and inefficient utilization of data in bi-level optimization also account for suboptimal results. We adopt a more realistic objective focusing on the performance of subnets and simplify it with the help of the information-based measurements. 
Finally, we explain theoretically why progressively shrinking the width of the supernet is necessary and reduce the approximation error of optimal weights in DARTS. 
Our proposed method, named IS-DARTS, comprehensively improves DARTS and resolves the aforementioned problems. Extensive experiments on NAS-Bench-201 and DARTS-based search space demonstrate the effectiveness of IS-DARTS.
\end{abstract}

\section{Introduction}

Although neural networks have achieved significant success in numerous computer vision domains\cite{he2016deep,ren2015faster,long2015fully}, designing neural architectures remains an uphill task demanding extensive expertise and repeated fine-tuning. In recent years, researchers have endeavored to discover Neural Architecture Search (NAS) methods that automate the manual process of exploiting outstanding architectures. Previous NAS methods based on reinforcement learning\cite{baker2016designing,zoph2018learning,pham2018efficient} and evolutionary algorithm\cite{real2017large,liu2017hierarchical,real2019regularized} typically require thousands of GPU days computational resources, which impede their practical application. Therefore, one-shot methods\cite{liu2018darts,chen2019progressive,chu2020fair,chu2020darts} are proposed, which complete searching in one training of the supernet containing all potential subnets with weight sharing strategy. 

Among many proposed approaches, Differential ARchitecture Search (DARTS)\cite{liu2018darts} leverages architecture parameters to continuously relax the discrete search space, enabling gradient backpropagation in the supernet. DARTS enhances the search efficiency by a large margin while achieving comparable performance. However, some studies\cite{chen2019progressive,zela2019understanding,chu2020fair} report that DARTS suffers from performance collapse due to the dominance of parameter-free operations in the selected architecture. Existing solutions to this problem vary considerably, including imposing handcrafted rules\cite{chu2020fair,chu2020darts}, suppressing the indicator\cite{liang2019darts+,zela2019understanding} and so on. However, the vast majority rely on the prior knowledge that performance collapse will happen, which cure the symptoms instead of the disease. Moreover, many methods lack theoretical analysis on their relationship with the original optimization problem, making them less persuasive. 

\begin{figure*}[t]
  \centering
  \includegraphics[width=\linewidth]{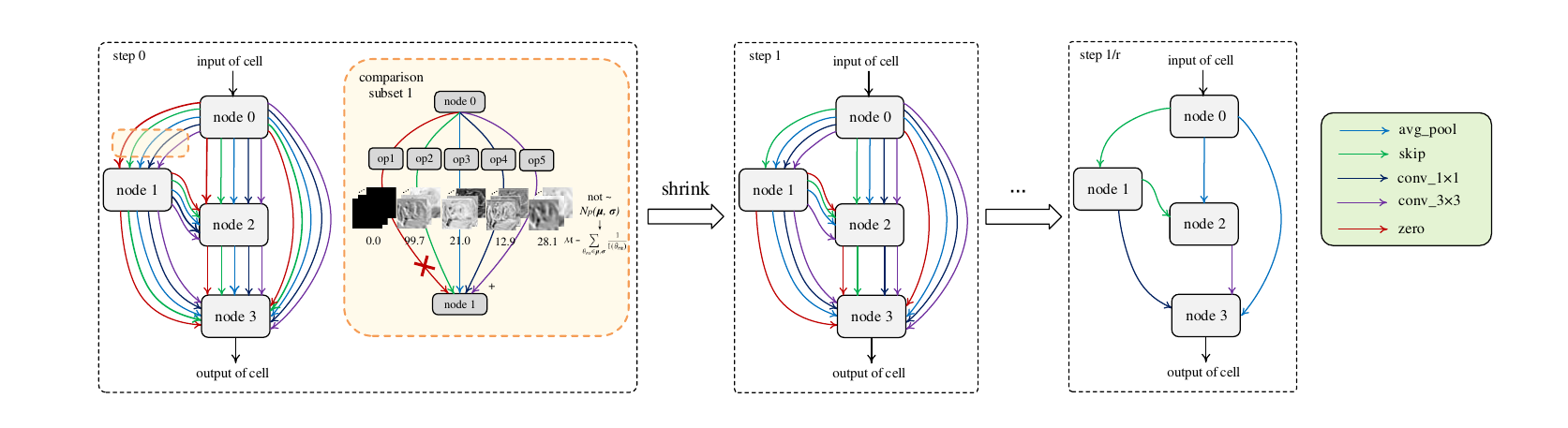}
  \caption{Overall illustration of IS-DARTS. The searching process of only one cell in the supernet is displayed for simplicity. The operations in one comparison subset(one edge for example in the figure) are compared and gradually discarded through the information-based metric.}
  \label{fig:illustration}
\end{figure*}

This paper is dedicated to analyzing and resolving the problems leading DARTS to suboptimal results, including but not limited to the above two problems. 
We prove theoretically and experimentally that the architecture parameters in the supernet fail to represent the importance of the candidate operations in the search space. As a result, the subnet selected by the magnitude of architecture parameters undisputedly performs disastrously. In response, we measure the importance more precisely through the information in the intermediate feature maps output by the operations. We also prove that nodes in a subnet retaining more important operations output more informative and significant feature maps.

Furthermore, we find that the losses minimized in in the bi-level optimization objective of DARTS are all concerned with the supernet. We reformulate the objective with binary masks and turn its attention to the performance of the subnets, which should be the primary objective of DARTS. In addition, the training set encounter a forcible split for the training of the network weights and the architecture parameters, which reduces the efficiency of data utilization. We emancipate the validation set in the objective function and calculate more precise weights with the help of the importance measurement. 

Last, the approximation in the objective introduces error into the training of the network weights, which in turn affects the identification of good subnets. We explain the necessity for introducing multi-step shrinking in order to decrease the approximation errors. Following previous researches\cite{chen2019progressive,li2020sgas,wang2021rethinking}, we explore to progressively shrink the operations in a defined comparison subset, which connects the supernet with the subnet step by step.

The proposed method, named Information-based Shrinking DARTS (IS-DARTS), is illustrated in Figure \ref{fig:illustration}. Extensive experiments on various search spaces and datasets manifest the great potential of the proposed IS-DARTS, which achieve better performance while costing less computational resources. On NAS-Bench-201, we achieve the same state-of-the-art results in 4 independent runs with only 2.0 GPU hours. The implementation of IS-DARTS is available at https://github.com/HY-HE/IS-DARTS.

\section{Related Works}

The efforts made to address the performance collapse of DARTS\cite{liu2018darts} can be roughly divided into two parts.

The first type is based on the searching supernet. 
Fair DARTS\cite{chu2020fair} utilizes sigmoid functions instead of softmax to convert the exclusive competition into a cooperative relationship. 
DARTS-\cite{chu2020darts} adds a decaying auxiliary skip connection apart from the original operations to suppress the potential advantage. 
iDARTS\cite{wang2021idarts} interposes a static BN layer between the node input and the operations to ensure norm consistency among different outputs. 
These methods make handcrafted rules and seek direct solutions, which may mistakenly reject good subnets.
SNAS\cite{xie2018snas}, DSNAS\cite{hu2020dsnas} and DrNAS\cite{chen2020drnas} introduce stochasticity and try different prior distributions of architecture parameters.
These methods encourage exploration in the search space to boost the likelihood of parameterized candidates occurring.
PC-DARTS\cite{xu2019pc} and DropNAS\cite{hong2022dropnas} presents dropout on feature map channels and grouped operations to prevent the supernet from overfitting, but parts of the supernet may lack sufficient training.

The second type is based on the searching pipeline. 
DARTS+\cite{liang2019darts+} and PDARTS\cite{chen2019progressive} limits the number of skip connections to a constant, which is forced and suspicious. 
RDARTS\cite{zela2019understanding} and SDARTS connects the performance collapse with the proliferation of the Hessian eigenvalues of the validation loss as an indicator. Such methods rely heavily on the quality of the indicator.
SP-DARTS\cite{zhang2021robustifying} and Beta-DARTS\cite{ye2022beta} adopts sparse regularized approximation and Beta-Decay regularization to adjust the distribution of the architecture parameters.
PDARTS\cite{chen2019progressive}, SGAS\cite{li2020sgas}, PT\cite{wang2021rethinking} and OPP-DARTS\cite{zhu2021operation} fill in the gap between the searching stage and the evaluation stage by progressively increasing the architecture depths or widths. These methods rely on emphasizing the defect of skip connections when drawing the supernet close to the selected subnet.

\section{Methods}

\subsection{Preliminaries}

In the searching stage of DARTS\cite{liu2018darts}, the supernet consists of normal cells for feature extraction and reduction cells for feature downsampling. In each cell, a direct acyclic graph with N sequential nodes is connected with edges. The $i$-th node is an intermediate feature map $x^{(i)}$, and the edge $(i,j)$ connecting the $i$-th node to the $j$-th node includes all the candidate operations $\{o_{k}^{(i,j)}|k\}$ in the search space $\mathcal{S}$. DARTS applies continuous relaxation through trainable architecture parameters $\alpha$ to weigh the outputs of operations. So the computation on the edge is:
\begin{equation}
  \bar{o}^{(i,j)}(x^{(i)})=\sum_{k=1}^{\vert\mathcal{S}\vert}\frac{exp(\alpha^{(i,j)}_{k})}{\sum_{k'=1}^{\vert\mathcal{S}\vert}exp(\alpha^{(i,j)}_{k'})}o_{k}^{(i,j)}(x^{(i)})
\end{equation}
where $\bar{o}^{(i,j)}$ is the output of edge $(i,j)$. Two sets of $\alpha=\{\alpha^{(i,j)}_k|i,j,k\}$ are respectively shared in all the normal and reduction cells. Node $x^{(j)}$ is the summarization of all the edges connected with the predecessors:
\begin{equation}
  x^{(j)}=\sum_{i<j}\bar{o}^{(i,j)}(x^{(i)})
\end{equation}
In this way, the discrete operation selection problem can be converted to a differentiable bi-level optimization objective:
\begin{equation}\label{eq3}
\begin{aligned}
  &\alpha^{*}=arg\min_{\alpha}\mathcal{L}_{val}(w^{*}(\alpha),\alpha)\\
  &s.t. w^{*}(\alpha)=arg\min_{w}\mathcal{L}_{train}(w,\alpha)
\end{aligned}
\end{equation}
Such objective assumes that the best network weights $ w^{*}(\alpha)$ corresponding to every set of architecture parameters $\alpha$ can be obtained, which is unpractical due to the continuity of $\alpha$.

\subsection{Rethinking the Defects in DARTS}

\begin{figure}[t]
  \centering
  \includegraphics[width=\linewidth]{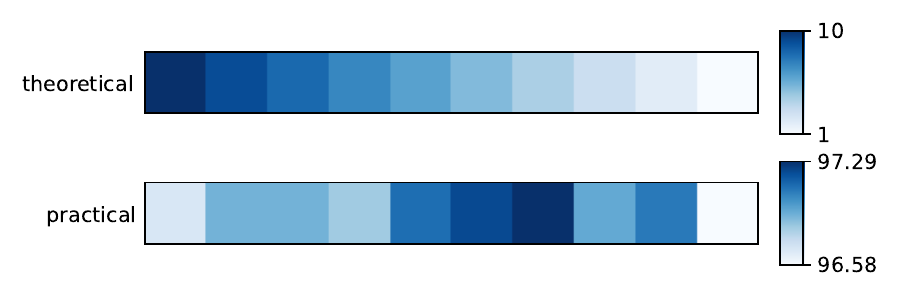}
  \caption{Experiments on operation importance represented by architecture parameters in DARTS. We always replace an operation with a less important one. The upper part shows the theoretical ranking of accuracies, while the lower part shows the practical evaluation accuracies of subnets.}
  \label{fig:importance}
\end{figure}

Several methods\cite{chu2020darts,liang2019darts+,zela2019understanding,chen2020stabilizing,chen2019progressive} mentioned in section 2 are derived from the prior knowledge that skip connections will dominate in DARTS. They aim to  alleviate the phenomenon rather than eliminate it, and therefore cannot fundamentally solve the performance collapse problem. \cite{chu2020darts} proposes that except for a candidate operation, they also functions as an auxiliary connection to stabilize the supernet training. As a result, the architecture parameters $\alpha$ of skip connections are unreasonably high and incommensurable with other operations. Further, an operation with a high architecture parameter merely plays an important role when all operations work together to find the correct labels. However, it is a completely different situation when the operation works alone in a subnet. $\alpha$’s preference to skip connections is a prominent example, considering that skip connections are of vital importance in modern CNNs\cite{he2016deep}. Thus, the root cause of the performance collapse is: \\
\textbf{Defect 1:} The architecture parameters $\alpha$ cannot represent the importance of operations in the search space.

To demonstrate it, we conduct an experiment on DARTS that consistently replaces a more important operation represented by a higher $\alpha$ with a less important operation represented by a lower $\alpha$ and compares the evaluation accuracies of the two subnets. Specifically, the operations are divided into the winners and the losers in a random run of DARTS. Every time, the most important operation identified by DARTS with the highest $\alpha$ among the winners are replaced with the operation with the highest $\alpha$ among the losers. Theoretically, the evaluation performance of the corresponding subnets should descend, shown in the upper part of Figure \ref{fig:importance}. However, the lower part displaying the obtained evaluation accuracies are unordered in fact. 
Fundamental improvements can only be achieved by exploring precise standards to represent the importance of operations.

In addition, although bi-level optimization has contributed to the success of DARTS, we find that it suffers from three crucial weaknesses: \\
\textbf{Defect 2:} The excessive concern over the supernet, the approximation of optimal weights and the inefficiency of data utilization in bi-level optimization can result in poor performance.

\begin{figure}[t]
  \centering
  \includegraphics[width=\linewidth]{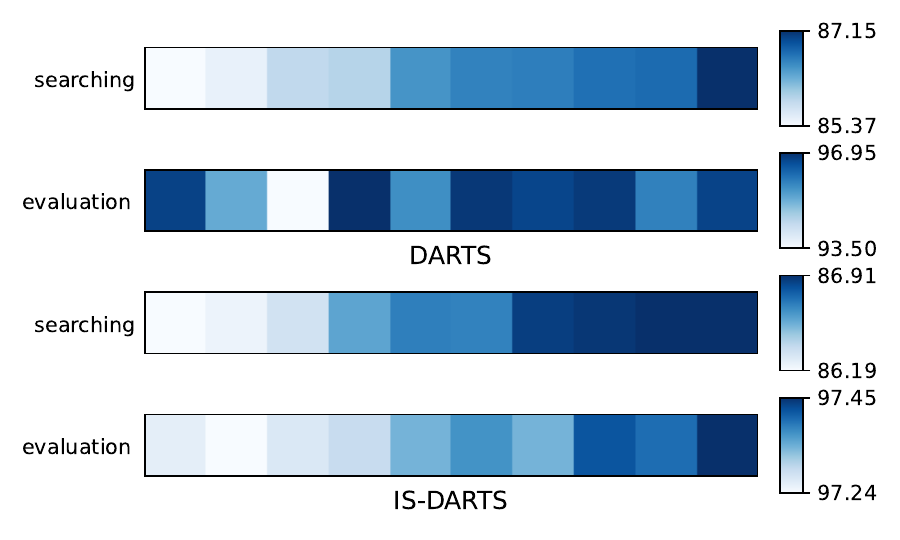}
  \caption{Comparison of the accuracies in 10 runs of DARTS and IS-DARTS. The upper part shows the searching stage while the lower shows the evaluation stage. The two color bars for DARTS have few linear correlation.}
  \label{fig:gap}
\end{figure}

First, the losses minimized in the bi-level optimization objective are all calculated on the supernet, leading to excessive concern over the performance of the supernet. It has no relation with the validation accuracies of the subnet. We compare the final accuracies of the search and evaluation stages in several runs of DARTS in Figure \ref{fig:gap}, where the two lists of accuracies have few linear correlation. Previous reserches\cite{chen2019progressive,li2020sgas,wang2021rethinking,wang2021rethinking,zhu2021operation} leverage progressive pipelines to connect the two stages, but lack theoretical analysis on why such designs is superior, which is unconvincing.

Second, unlike the original objective, the actual training process optimizes the architecture parameters and network weights alternatively. The network weights are updated with only one gradient descent under fixed architecture parameters, and are served as an approximation of the optimal weights, which introduces errors and makes the results sub-optimal:
\begin{equation}
\begin{aligned}
  &\nabla _{\alpha}\mathcal{L}_{val}(w^{*}(\alpha),\alpha)\\
  \approx&\nabla _{\alpha}\mathcal{L}_{val}(w-\xi \nabla _{w}\mathcal{L}_{train}(w,\alpha),\alpha)
\end{aligned}
\end{equation}

Third, training of both architecture parameters and network weights requires samples, which forces DARTS to divide the training set into two halves. Consequently, both trainable parameters are shown only a portion of the samples, which reduces the utilization of data and aggravates performance. 

\subsection{Precise Measurement via Fisher Information}

To find the precise measurement of the importance of candidate operations, we start by assuming that after a significant and informative output feature map (FM) $X'\in \mathbb{R}^{H\times W}$ is flattened to $X\in \mathbb{R}^{p}$, it can be estimated with a $p$-variate random distribution $\mathcal{D}(\bm{\theta})$, where $H$,$W$ are the height and width of the feature map respectively, $p=H*W$ is the total number of pixels, and $\bm{\theta}$ is the vector of $\mathcal{D}$ parameters.

Let $\bm{\theta}(\chi)$ be an estimator that indicate how well $\mathcal{D}(\bm{\theta})$ fits X based on the observed FMs $\chi$. For example, a constant function that maps to a specific $\bm{\theta}$ regardless of $\chi$ is undoubtedly a bad estimator. Thus mean square error (MSE) is introduced to evaluate the quality of the estimator. If the optimal parameter is $\bm{\theta}^{*}$, MSE is defined as:
\begin{equation}
  L_{MSE}=\operatorname{E}[(\bm{\theta}(\chi)-\bm{\theta}^{*})^{T}(\bm{\theta}(\chi)-\bm{\theta}^{*})]
\end{equation}
MSE can be decomposed as:
\begin{equation}
  L_{MSE}=\operatorname{tr}(\operatorname{Cov}[\bm{\theta}(\chi)])+(\operatorname{E}[\bm{\theta}(\chi)]-\bm{\theta}^{*})^{T}(\operatorname{E}[\bm{\theta}(\chi)]-\bm{\theta}^{*})
\end{equation}
where $\operatorname{tr}(\cdot)$ is the trace of matrix. If the estimator is restricted to be unbiased, i.e., $\operatorname{E}[\bm{\theta}(\chi)]-\bm{\theta}^{*}=\bm{0}$, MSE becomes exactly the trace of the covariance, and the estimator with the smallest values on the diagonal is the best. According to Cramer–Rao inequality, under regularity conditions, the lower bound of MSE can be calculated by:
\begin{equation}\label{eq7}
  \operatorname{tr}(\operatorname{Cov}[\bm{\theta}(\chi)]) \ge \sum_{m}^{p}\frac{1}{\operatorname{I}(\theta_{m})}
\end{equation}
where $\operatorname{I}(\cdot)$ is the Fisher information, which is defined as:
\begin{equation}\label{eq8}
\begin{aligned}
  \operatorname{I}(\theta_m)&=\operatorname{E}\left[\left(\frac{\partial}{\partial\theta_m} \log f(X;\bm{\theta})\right)^2\right] \\
  &=-\operatorname{E}\left[\frac{\partial^2}{\partial\theta_m^2} \log f(X;\bm{\theta})\right]
\end{aligned}
\end{equation}
where $f(X;\bm{\theta})$ is the probability density function (PDF). 

Based on the above inference, the critical distribution $\mathcal{D}$ should be precisely determined since FMs vary with different inputs and depths in networks. However, we can in turn distinguish which FMs are insignificant by measuring whether they are close to interfering noise. Therefore, we assume that all the pixels are normally distributed and independent, i.e., $X\sim N_{p}(\bm{\mu},\bm{\Sigma})$, thus PDF of $X$ can be simplified as:
\begin{equation}
  f(X)=\frac{1}{\sqrt{2\pi}^{p} \prod_{m}^{p}\sigma_{m}} \operatorname{exp}\left(\sum_{m}^{p} -\frac{(x_{m}-\mu_{m})^{2}}{2\sigma_{m}^{2}} \right)
\end{equation}
where $\bm{\Sigma}=\operatorname{diag}(\bm{\sigma})$. The parameters $\bm{\theta}$ become $\{\bm{\mu},\bm{\sigma}\}=\{\mu_m,\sigma_m|m\}$. According to Equation (\ref{eq8}), the Fisher information with respect to $\mu_m$ and $\sigma_m$ are:
\begin{equation}
\begin{aligned}
  &\operatorname{I}(\mu_m)=\operatorname{E}\left[\frac{1}{\sigma_m^{2}}\right]\\
  &\operatorname{I}(\sigma_m)=\operatorname{E}\left[\frac{3(x_m-\mu_m)^{2}}{\sigma_m^{4}}-\frac{1}{\sigma_m^{2}}\right]
\end{aligned}
\end{equation}
A lower bound of MSE when $\mathcal{D}$ is $N_{p}$ is obtained by substituting into the last term of Equation (\ref{eq7}), defined as the Information-based Importance Measurement (IIM) $\mathcal{M}$ of FMs:
\begin{equation}\label{eq12}
  \mathcal{M}=\sum_{\theta_m \in {\bm{\mu},\bm{\sigma}}}\frac{1}{\operatorname{I}(\theta_m)}=\sum_{m}^{p}\frac{\sigma_m^{2}}{3}
\end{equation}
where the variance $\sigma_m$ is calculated across channels. When $\mathcal{M}$ increases, the lower bound of MSE increases, $N_{p}$ is a worse estimator, and $\chi$ are less likely to be nonsensical noises. An interesting fact is that the Fisher information matrix of $\bm{\theta}$ is a $2p \times 2p$ matrix and has typical element:
\begin{equation}
  (\operatorname{I}(\bm{\theta}))_{u,v}=\operatorname{E}\left[\left(\frac{\partial}{\partial\theta_u} \log f(X;\theta_u)\right)\left(\frac{\partial}{\partial\theta_v} \log f(X;\theta_v)\right)\right]
\end{equation}
It is easy to prove that $\forall u\ne v,(\operatorname{I}(\bm{\theta}))_{u,v}=0$, thus the Fisher information matrix of $\bm{\theta}$ is a diagonal matrix.
As a result, $\mathcal{M}$ is exactly the inverse of the sum of the Fisher information matrix. When $\mathcal{M}$ increases, the peak around the maximum value of the log-likelihood function is shallow. Therefore from another perspective, $\chi$ provides little information about the certainty of the parameters of the normal distribution in the estimation process and furthermore are less likely to be noises.

We can now measure the importance of operations via their output FMs, but what is the meaning of this importance for the supernet? We try to answer this question by broadening our vision from operations to nodes.

Consider the situation that the weights of the operations in the supernet are optimal after the complete training process. The intermediate FM in a node $j$ is the sum of FMs produced by operations on edges $\{(i,j)|i\}$. Considering that operations are calculated separately in the forward propagation, we assume that the FM $X_s$ in a node is the sum of $q$ independent $p$-variate normal distribution, which is proved to obey a normal distribution as well according to characteristic functions:
\begin{equation}
  X_s=\sum_{n}^{q}X_{n}\sim N_{p}(\bm{\mu}_s,\bm{\Sigma}_{s})
\end{equation}
where $q=|\{(i,j)|i\}|*|\mathcal{S}|$ is the number of candidate operations connected to node $j$, and $\bm{\mu}_s=\sum_{n}^{q}\bm{\mu}_n,\bm{\Sigma}_s=\sum_{n}^{q}\bm{\Sigma}_{n}$ are the sum of means and variances respectively.

According to Equation (\ref{eq8}), the Fisher Informations with respect to the mean and the variance of the $m$-th element in the $n$-th distribution are:
\begin{equation}
\begin{aligned}
  &\operatorname{I}(\mu_{ms})=\operatorname{E}\left[\frac{1}{\sum_{n}^{q}\sigma_{mn}^{2}}\right]\\
  &\operatorname{I}(\sigma_{ms})=\operatorname{E}\left[\frac{3(x_{ms}-\sum_{n}^{q}\mu_{mn})^{2}}{\left(\sum_{n}^{q}\sigma_{mn}^{2}\right)^{2}}-\frac{1}{\sum_{n}^{q}\sigma_{mn}^{2}}\right]
\end{aligned}
\end{equation}
where  $\sigma_{mn}$ is the $m$-th element of the main diagonal of $\bm{\Sigma}_n$. According to Equation (\ref{eq12}), the IIM $\mathcal{M}_s$ of the FM in the node is:
\begin{equation}
  \mathcal{M}_s=\sum_{m}^{p}\frac{\sum_{n}^{q}\sigma_{mn}^{2}}{3}=\sum_{n}^{q}\mathcal{M}_n
\end{equation}

When a constant number of operations are left in this node, they output the most significant $X_n$s with the maximal measurements $\mathcal{M}_n$s, and therefore the most significant $X_s$s with the maximal measurement $\mathcal{M}_s$. Briefly, keeping the operations with the highest IIMs in the subnet means retaining the highest IIM for the node compared to keeping the same number of other operations. 

\begin{algorithm}[t]
\caption{Pipeline of IS-DARTS}
\label{alg:alg1}
\textbf{Input}: search space $\mathcal{S}$, shrink rate $r$\\
\textbf{Output}: mask $\Gamma$
\begin{algorithmic}[1] 
\STATE initialization: mask $\Gamma^{*}_{0}\leftarrow\bm{1}^{|\{(i,j)\}|*q}$, $\forall h$, reserved subset $A^{h}_{0}\leftarrow O^h$, discarded subset $B^{h}_{0}\leftarrow \emptyset$
\STATE construct supernet on $\mathcal{S}$
\FOR{$z$ in $[1,1/r]$}
\WHILE{not converged}
\STATE train the subnet under mask $\Gamma^{*}_{z-1}$ with training set
\ENDWHILE
\STATE obtain the optimal weights $w^{*}(\Gamma^{*}_{z-1})$
\STATE compute $\mathcal{M}_{z}$ of operations in $A_{z}$ averaged over part of validation set
\FOR{all h}
\STATE seperate $\mathcal{M}^h_{z}$ corresponding to $h$-th comparison subset
\STATE rank $\mathcal{M}^h_{z}$ in an increasing order as $\hat{\mathcal{M}}^{h}_{z}$
\STATE move $D^{h}_{z}$ from $A^{h}_{z-1}$ to $B^{h}_{z-1}$ corresponding to the top $r(|\phi^h|-C)$ values in $\mathcal{M}^h_{z}$
\STATE update mask $\Gamma^{*}_{z-1}$ corresponding to operations in $D^{h}_{z}$ to 0 to get $\Gamma^{*}_{z}$
\ENDFOR
\ENDFOR
\STATE \textbf{return} $\Gamma^{*}_{1/r}$
\end{algorithmic}
\end{algorithm}

\subsection{Improvements on Bi-level Optimization}

DARTS applies the bi-level optimization due to the existence of architecture parameters $\alpha$ and is demonstrated to suffer from crucial weaknesses in section 3.2. With IIM, we no longer rely on $\alpha$ to select candidate operations in the training of the supernet. Instead, we adopt a binary mask $\Gamma\in\mathbb{R}^{|\{(i,j)\}|*q}$ to represent the presence of candidates, where $|\{(i,j)\}|$ is the number of edges in a cell. Hence the objective optimization changes to:
\begin{equation}\label{eq16}
\begin{aligned}
  &\Gamma^{*}=arg\min_{\Gamma}\mathcal{L}_{val}(w^{*}(\Gamma),\Gamma)\\
  &s.t. w^{*}(\Gamma)=arg\min_{w}\mathcal{L}_{train}(w,\Gamma),\forall h,\left \| \Gamma^{h} \right \|=C
\end{aligned}
\end{equation}
where $\left \| \Gamma^{h} \right \|$ and $C$ are determined by the requirement of the search space. 
In the supernet, different operations are divided into groups and compared in a winner-take-all game that selects the best $C$ candidates. We name such a group the comparison subset $\phi$, then $\cup_{h}\phi^{h}=\{o^{(i,j)}_{k}|i,j,k\}, \cap^{h}\phi^{h}=\emptyset$. For example, DARTS-based search space requires that two are selected among all the operations input to a certain node, i.e. $\phi^{h}=\{o^{(i,h)}_{k}|i,k\}$, so the maximum of $h$ is the number of nodes and $C=2$. $\Gamma ^{h}$ is the mask corresponding to the $h$-th comparison subset $\phi^{h}$, therefore the concentration of the masks $\operatorname{concat}_h(\Gamma^{h})=\Gamma$. $\left \| \Gamma ^{h} \right \|$ represents the L1 norm of $\Gamma ^{h}$.

Compared to Equation (\ref{eq3}), Equation (\ref{eq16}) is advantageous in that it focuses on the validation performance of the subnet with $\Gamma$ covering the eliminated candidates, rather than the supernet, which is supposed to be the primary problem of DARTS.

Though $\Gamma$ is a discrete value, both the objective function of Equation (\ref{eq16}) that enumerate all the optimal subnets on the training set corresponding to each $\Gamma$ and the constraint condition that evaluate all the subnets on the validation set is not computationally feasible. We simplify the proposed objective by approximating $\mathcal{L}_{val}$:
\begin{equation}\label{eq17}
\begin{aligned}
  &\Gamma^{*}=arg\min_{\Gamma}{\Delta}_{IIM}(w^{*}(\Gamma),\Gamma)\\
  &s.t. w^{*}(\Gamma)=arg\min_{w}\mathcal{L}_{train+val}(w,\Gamma),\forall h,\left \| \Gamma^{h} \right \|=C
\end{aligned}
\end{equation}
where ${\Delta}_{IIM}$ is defined as the decreased IIM comparing the supernet to the subnet masked by $\Gamma$. Thus the selection of $\Gamma^{*}$ needs no validation, and all samples can be used to train a more precise $w^{*}$.

As for the constraint condition, we further simplify Equation (\ref{eq17}) to the one-step selection form:
\begin{equation}
\begin{aligned}
  \Gamma^{*}&=arg\min_{\Gamma}{\Delta}_{IIM}(w^{*}(\Gamma),\Gamma)\\
  s.t. &w^{*}(\Gamma)\approx w^{*}(\Gamma_{0})=arg\min_{w}\mathcal{L}_{train+val}(w,\Gamma_{0}),\\
  &\forall h,\left \| \Gamma^{h} \right \|=C
\end{aligned}
\end{equation}
where $\Gamma_{0}=\bm{1}^{|\{(i,j)\}|*q}$ corresponds to the initial supernet. Such optimization assumes that the weights in the supernet are still optimal in the subnet under $\Gamma$, which obviously introduces errors.

\begin{table*}[t]
  \caption{Comparison with state-of-the-art methods under NAS-Bench-201 search space. The accuracies are obtained on 4 independent runs.}
  \label{tab:bench}
  \small
  \resizebox{\linewidth}{!}{
  \begin{tabular}{cccccccc}
    \toprule
        \multirow{2}{*}{Method} & \multirow{2}{*}{Cost(hours)} & \multicolumn{2}{c}{CIFAR-10} & \multicolumn{2}{c}{CIFAR-100} & \multicolumn{2}{c}{ImageNet-6-120} \\ \cline{3-8}
        ~ & ~ & valid & test & valid & test & valid & test \\
    \midrule
        ENAS\cite{pham2018efficient} & 3.7 & 37.51±3.19 & 53.89±0.58 & 13.37±2.35 & 13.96±2.33 & 15.06±1.95 & 14.84±2.10 \\
        DARTS(1st)\cite{liu2018darts} & 3.2 & 39.77±0.00 & 54.30±0.00 & 15.03±0.00 & 15.61±0.00 & 16.43±0.00 & 16.32±0.00 \\
        DARTS(2nd)\cite{liu2018darts} & 10.2 & 39.77±0.00 & 54.30±0.00 & 15.03±0.00 & 15.61±0.00 & 16.43±0.00 & 16.32±0.00 \\
        GDAS\cite{dong2019searching} & 8.7 & 89.89±0.08 & 93.61±0.09 & 71.34±0.04 & 70.70±0.30 & 41.59±1.33 & 41.71±0.98 \\
        SETN\cite{dong2019one} & 9.5 & 84.04±0.28 & 87.64±0.00 & 58.86±0.06 & 59.05±0.24 & 33.06±0.02 & 32.52±0.21 \\
        DSNAS\cite{hu2020dsnas} & - & 89.66±0.29 & 93.08±0.13 & 30.87±16.40 & 31.01±16.38 & 40.61±0.09 & 41.07±0.09 \\
        PC-DARTS\cite{xu2019pc} & - & 89.96±0.15 & 93.41±0.30 & 67.12±0.39 & 67.48±0.89 & 40.83±0.08 & 41.31±0.22 \\
        DARTS-\cite{chu2020darts} & 3.2 & 91.03±0.44 & 93.80±0.40 & 71.36±1.51 & 71.53±1.51 & 44.87±1.46 & 45.12±0.82 \\ \hline
        IS-DARTS & 2.0 & 91.55±0.00 & 94.36±0.00 & 73.49±0.00 & 73.51±0.00 & 46.37±0.00 & 46.34±0.00 \\ 
  \bottomrule
  \end{tabular}
  }
\end{table*}

\subsection{Necessity of Shrinking Supernet}

As mentioned above, direct one-step selection leaps from the supernet to the subnet and leads to suboptimal results. Specifically, when an operation is removed, the operations at the exact location in all the cells are removed, influencing all the intermediate FMs. In addition, the weights in the operations are no longer optimal under the contractive architecture. As a result, we need to progressively discard the operations connected to the node, followed by retraining the contractive architecture and re-measuring the importance.

We apply a greedy strategy that forces $\left \| \Gamma^{h} \right \|$ to step towards $C$ gradually. The final optimization objective is listed as:
\begin{equation}
\begin{aligned}
  \Gamma^{*}_{z}&=arg\min_{\Gamma_{z}}\Delta_{IIM}(w^{*}(\Gamma_{z}),\Gamma_{z})\\
  s.t. &w^{*}(\Gamma_{z})\approx w^{*}(\Gamma^{*}_{z-1})=arg\min_{w}\mathcal{L}_{train+val}(w,\Gamma^{*}_{z-1}),\\
  &\forall h,\left \| \Gamma^{h}_{z} \right \|=C^{h}_{z}
\end{aligned}
\end{equation}
where z is the count of steps. The number of operations reserved in the comparison subset $\phi^h$ is defined as $C^{h}_{z}=|\phi^h|-zr(|\phi^h|-C)$, where $r$ is the ratio at which operations are discarded at one step. Therefore, $C_0^{h}=|\phi^h|,C_{1/r}^{h}=C$, which connects the searching and evaluation stage. Such optimization assumes that the weights in the subnet under the mask $\Gamma^{*}_{z-1}$ are still optimal in the subnet under the mask $\Gamma_{z}$, though a few operations are discarded. Condition $\left \| \Gamma^{h}_{z} \right \|=C^{h}_{z}$ equals to finding two subsets of $O^h$, $A^{h}_{z}$ to be reserved in the subnet and $B^{h}_{z}$ to be discarded:
\begin{equation}
\begin{aligned}
  A^{h}_{z}=\phi^{h}\otimes \Gamma^{h}_{z}&=\{o^{h}_{a^{h}_{l}}|l\in[1,C^{h}_{z}]\}\\
  B^{h}_{z}=\phi^{h}-\phi^{h}\otimes \Gamma^{h}_{z}&=\{o^{h}_{b^{h}_{l}}|l\in[1,|\phi^h|-C^{h}_{z}]\}
\end{aligned}
\end{equation}
where $\otimes$ means element-wise multiplication and then take out zero values, $a^{h}_{l},b^{h}_{l}$ are the indices of the reserved and discarded operations in $\phi^{h}$ respectively. $\forall h,z,A^{h}_{z}\cup B^{h}_{z}=\phi^h,A^{h}_{z}\cap B^{h}_{z}=\emptyset$.

To realize the algorithm, we first train the subnet with mask $\Gamma^{*}_{z-1}$ in the previous step to the optimal. Note that except for the first step that starts with a randomly initialized supernet, since $\Gamma^{*}_{z}$ changes little compared to $\Gamma^{*}_{z-1}$, the optimal weights $w^{*}(\Gamma^{*}_{z})$ can be obtained through a few epochs of training with little error based on $w^{*}(\Gamma^{*}_{z-1})$, which greatly reduces computational resources. Then The IIMs of FMs output by $A^{h}_{z-1}$ is calculated as:
\begin{equation}
  \mathcal{M}^{h}_{z}=\{\mathcal{M}^{h}_{a^h_l}|l\in[1,C^{h}_{z-1}]\}
\end{equation}
Subsequently, the IIMs are ranked in an increasing order:
\begin{equation}
\begin{aligned}
  \hat{\mathcal{M}}^{h}_{z}=\{\mathcal{M}^{h}_{d^h_l}|l\in[1,C^{h}_{z-1}]\}\\
  s.t.\forall l_{1}>l_{2},\mathcal{M}^{h}_{d^{h}_{l_{1}}}>\mathcal{M}^{h}_{d^{h}_{l_{2}}}
\end{aligned}
\end{equation}
where $d^h_l$ is the index of the $l$-th highest IIM. Finally, we take the first $r(|\phi^h|-C)$ values in $\hat{\mathcal{M}}^{h}_{z}$ and map them to the corresponding operations:
\begin{equation}
  D^{h}_{z}=\{o^{h}_{d^{h}_{l}}|l\in[1,r(|\phi^h|-C)]\}\subseteq A^{h}_{z-1}
\end{equation}
$D^{h}_{z}$ is moved from $A^{h}_{z-1}$ to $B^{h}_{z-1}$ as $A^{h}_{z}$ and $B^{h}_{z}$.

The whole algorithm is listed in Algorithm~\ref{alg:alg1}.

\section{Experiments}

All our experiments are implemented with Pytorch\cite{paszke2017automatic} and conducted on NVIDIA GTX 3090 GPUs.

\subsection{Results on NAS-Bench-201 Search Space}

\begin{figure*}[h]
  \centering
  \includegraphics[width=\linewidth]{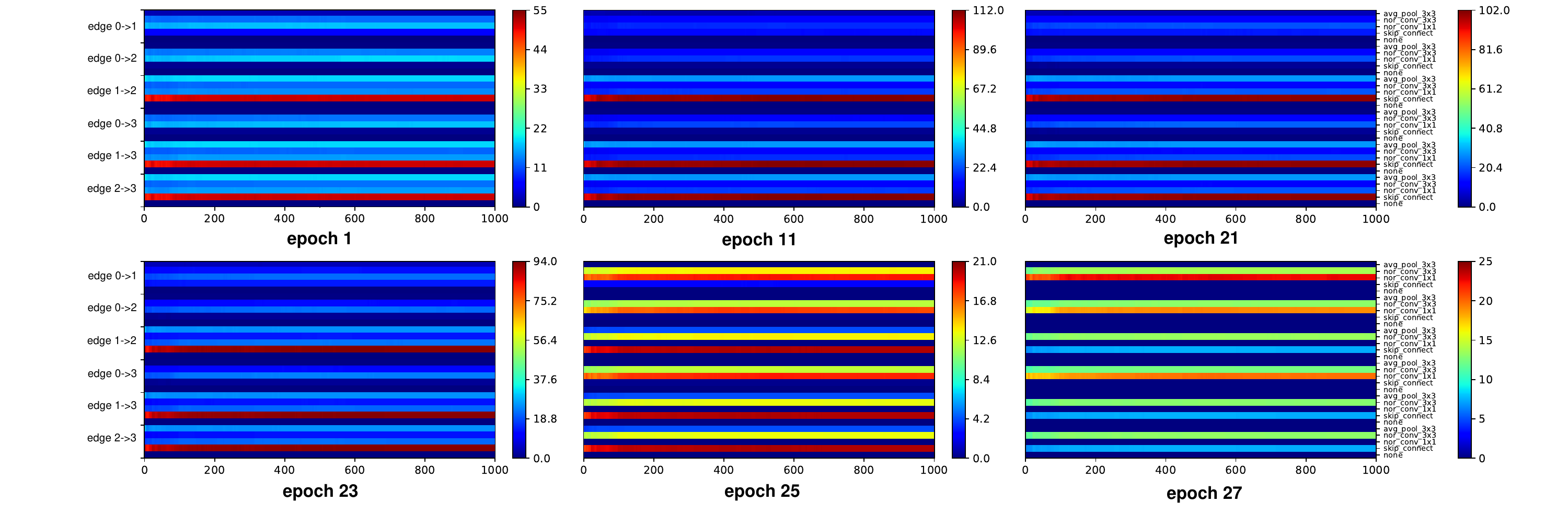}
  \caption{The IIMs of operations on all edges at different epochs in the searching stage. The x-axis represents the number of input images, and the y-axis represents operations at different locations in the cell. The mapping relation between colors and IIMs is shown in the right bars.}
  \label{fig:visualization}
\end{figure*}

\newcommand{\tabincell}[2]{\begin{tabular}{@{}#1@{}}#2\end{tabular}}  

\begin{table}[t]
  \caption{Comparison with state-of-the-art methods under DARTS search space on CIFAR-10. The accuracies are obtained on 4 independent runs.}
  \label{tab:cifar}
  \small
  \resizebox{\linewidth}{!}{
  \begin{tabular}{ccccc}
    \toprule
        Architecture & \tabincell{c}{ Test Error\\(\%)} & \tabincell{c}{ Params\\(M)} & \tabincell{c}{ Cost\\(GPU Days)} & \tabincell{c}{ Search\\Method} \\ 
    \midrule
        DenseNet-BC\cite{huang2017densely} & 3.46 & 25.6 & - & manual \\ 
        NASNet-A\cite{zoph2018learning} & 2.65 & 3.3 & 2000 & RL \\ 
        AmoebaNet-A\cite{real2019regularized} & 3.34±0.06 & 3.2 & 3150 & EA \\ 
        AmoebaNet-B\cite{real2019regularized} & 2.55±0.05 & 2.8 & 3150 & EA \\ 
        Hierarchical-EAS\cite{liu2017hierarchical} & 3.75±0.12 & 15.7 & 300 & EA \\ 
        PNAS\cite{liu2018progressive} & 3.41±0.09 & 3.2 & 225 & SMBO \\ 
        ENAS\cite{pham2018efficient} & 2.89 & 4.6 & 0.5 & RL \\ 
        DARTS(1st)\cite{liu2018darts} & 3.00±0.14 & 3.3 & 1.5 & gradient \\ 
        DARTS(2nd)\cite{liu2018darts} & 2.76±0.09 & 3.3 & 4 & gradient \\ 
        SNAS\cite{xie2018snas} & 2.85±0.02 & 2.8 & 1.5 & gradient \\ 
        GDAS\cite{dong2019searching} & 2.82 & 2.5 & 0.17 & gradient \\ 
        SGAS\cite{li2020sgas} & 2.67±0.21 & 3.9 & 0.25 & gradient \\ 
        P-DARTS\cite{chen2019progressive} & 2.50 & 3.4 & 0.3 & gradient \\ 
        PC-DARTS\cite{xu2019pc} & 2.57±0.07 & 3.6 & 0.1 & gradient \\ 
        FairDARTS\cite{chu2020fair} & 2.54±0.05 & 3.32±0.46 & - & gradient \\ 
        SDARTS-ADV\cite{chen2020stabilizing} & 2.61±0.02 & 3.3 & 1.3 & gradient \\ 
        DARTS-\cite{chu2020darts} & 2.59±0.08 & 3.5±0.13 & 0.4 & gradient \\ \hline
        IS-DARTS & 2.56±0.04 & 4.25±0.22 & 0.42 & gradient \\ 
    \bottomrule
  \end{tabular}
  }
\end{table}

\textbf{Settings.} As the most widely used NAS benchmark, NAS-Bench-201\cite{dong2020bench} provides the performance of 15,625 architectures on three datasets(CIFAR-10, CIFAR-100 and ImageNet). The comparison subset in NAS-Bench-201 is operations on a certain edge and $C=1$. We keep the searching and evaluation settings same as those of DARTS in \cite{dong2020bench}, except that the searching epoch is decreased to 30 in order to reduce consumption. Specific architectures and hyper-parameters are shown in Appendix A.1. 
The shrink rate $r$ in IS-DARTS is 0.25, and the epoch interval between steps is 2. 200 samples of the validation set are used to calculate a precise and stable IIM, which will be proved in section 4.3.

\textbf{Results.} The mean and standard deviation of 4 independent runs with different random seeds of IS-DARTS is presented in Table \ref{tab:bench}. The architecture found on CIFAR-10 achieves consistent state-of-the-art performance on all the three datasets, showing the superiority and generalization of IS-DARTS. It significantly outperforms DARTS by 113.43\% on average by tackling the performance collapse problem. In addition, since IS-DARTS only searches on CIFAR-10 with fewer epochs and requires few extra computational resources, it reduces the searching cost by 37.5\% compared to DARTS.

\subsection{Results on DARTS Search Space}

\textbf{Settings.} Classical DARTS search space\cite{liu2018darts} is another important benchmark for evaluating NAS methods. In the searching stage, the architecture is selected under CIFAR-10 dataset. The comparison subset in the DARTS search space is all the operations input into a certain node and $C=2$. Then in the evaluation stage, the architecture is evaluated on CIFAR-10 and transferred to ImageNet. All the searching and evaluation settings are kept the same as those of DARTS\cite{liu2018darts} in Appendix A.2
. The shrink rate in IS-DARTS is 0.143 and the epoch interval between steps is 3. 600 samples of the validation set are used to calculate the IIM.

\textbf{Results.} 
Table \ref{tab:cifar} summarizes the performance of IS-DARTS on CIFAR-10 compared to handcrafted architectures and NAS methods. With extremely low computational costs in the searching stage, the average test accuracy of 4 independent runs is 97.44±0.04\%, basically same as state-of-the-art methods. To verify the generalization ability, architecture found on CIFAR-10 can still ensure a state-of-the-art performance, namely 75.9\% on ImageNet. IS-DARTS surpasses all the methods that search on CIFAR-10, and is comparable to the methods that search directly on ImageNet and take more than $7\times$ GPU days. The impressive results demonstrate the effectiveness and efficiency of IS-DARTS. 

\begin{table}[t]
  \caption{Comparison with state-of-the-art methods under DARTS search space on ImageNet. The dataset on which the architecture is searched is shown behind the methods. I represents ImageNet and C represents CIFAR-10.}
  \label{tab:imagenet}
  \resizebox{\linewidth}{!}{
  \begin{tabular}{cccccc}
    \toprule
        \multirow{2}{*}{Architecture} & \multicolumn{2}{c}{Test Error(\%)} & \multirow{2}{*}{\tabincell{c}{ Params\\(M)}} & \multirow{2}{*}{\tabincell{c}{ Cost\\(GPU Days)}} & \multirow{2}{*}{\tabincell{c}{ Search\\Method}} \\ 
        ~ & top-1 & top-5 & ~ & ~ & ~ \\
    \midrule
        Inception-v1\cite{szegedy2015going} & 30.2 & 10.1 & 6.6 & - & manual \\ 
        MobileNet\cite{howard2017mobilenets} & 29.4 & 10.5 & 4.2 & - & manual \\ 
        ShuffleNet 2$\times$\cite{zhang2018shufflenet} & 26.3 & - & $\sim$5 & - & manual \\ 
        NASNet-A\cite{zoph2018learning} & 26 & 8.4 & 5.3 & 2000 & RL \\ 
        NASNet-B\cite{zoph2018learning} & 27.2 & 8.7 & 5.3 & 2000 & RL \\ 
        NASNet-C\cite{zoph2018learning} & 27.5 & 9 & 4.9 & 2000 & RL \\ 
        AmoebaNet-A\cite{real2019regularized} & 25.5 & 8 & 5.1 & 3150 & EA \\ 
        AmoebaNet-B\cite{real2019regularized} & 26 & 8.5 & 5.3 & 3150 & EA \\ 
        AmoebaNet-C\cite{real2019regularized} & 24.3 & 7.6 & 6.4 & 3150 & EA \\ 
        PNAS\cite{liu2018progressive} & 25.8 & 8.1 & 5.1 & $\sim$225 & SMBO \\ 
        DARTS\cite{liu2018darts}(C) & 26.7 & 8.7 & 4.7 & 4 & gradient \\ 
        SNAS\cite{xie2018snas}(C) & 27.3 & 9.2 & 4.3 & 1.5 & gradient \\ 
        GDAS\cite{dong2019searching}(C) & 27.5 & 9.1 & 4.4 & 0.17 & gradient \\ 
        SGAS\cite{li2020sgas}(C) & 24.38 & 7.39 & 5.4 & 0.25 & gradient \\ 
        P-DARTS\cite{chen2019progressive}(C) & 24.4 & 7.4 & 4.9 & 0.3 & gradient \\ 
        PC-DARTS\cite{xu2019pc}(I) & 24.2 & 7.3 & 5.3 & 3.8 & gradient \\ 
        FairDARTS\cite{chu2020fair}(I) & 24.4 & 7.4 & 4.3 & 3 & gradient \\ 
        SDARTS-ADV\cite{chen2020stabilizing}(C) & 25.2 & 7.8 & - & - & gradient \\ 
        DARTS-\cite{chu2020darts}(I) & 23.8 & 7 & 4.9 & 4.5 & gradient \\ \hline
        IS-DARTS(C) & 24.1 & 7.1 & 6.4 & 0.42 & gradient \\ 
    \bottomrule
  \end{tabular}
  }
\end{table}

\subsection{Stability}

To further understand the mechanism of IS-DARTS, we visualize the IIMs in the searching process under NAS-Bench-201 as the number of the input images randomly sampled from CIFAR-10 increase from 1 to 1000 in Figure \ref{fig:visualization}. In the beginning of the searching stage, the steady rise of the information means the network is learning. However, the decline appears at around epoch 20, which would be explained by overfitting. It is evident that skip connections are crucial in the training of the supernet, far beyond other operations at epoch 11. After two shrinking steps, i.e. epoch 25 and 27, several operations in the search space are discarded, and therefore the information of the skip connections decreases sharply without their assistance. Convolutional operations start to show their strengths and finally contribute to the excellent performance of the selected subnet in IS-DARTS. In addition, the IIM metric shows its efficiency that it stabilizes with conservatively 200 inputs and costs few extra resources.

\section{Conclusion}

In this paper, we propose Information-based Shrinking DARTS(IS-DARTS). We precisely measure the importance of candidate operations in the search space via Fisher Information metric, and reformulate the bi-level optimization problem into a more accurate and efficient form focusing on the performance of subnets. Furthermore, we adopt a shrinking pipeline on the comparison subsets of the supernet. The remarkable performance of IS-DARTS demonstrates its ability to search for stable and outstanding neural architectures and apply in practical deep learning problems.

\section{Acknowledgments}
This work was supported by the National Basic Strengthen Research Program of ReRAM under Grant2022-00-03, Natural Science Foundation of ShaanxiProvince, China under Grant 2022JM-366 and by the Open-End Fund of Beijing Institute of Control Engineering under Grant OBCandETL-2022-03.

\bibliography{aaai24}

\clearpage

\appendix
\renewcommand{\thetable}{A.\arabic{table}}
\renewcommand{\thefigure}{A.\arabic{figure}}
\setcounter{figure}{0}
\setcounter{table}{0}

\centerline{{\LARGE \textbf{Supplementary Material}}}

\section{Training Settings}

\subsection{NAS-Bench-201 search space}

The supernet is stacked by $N=5$ cells in each of the 3 blocks and each cell is constructed by $V=4$ nodes connected with $E=6$ edges. 
Each edge contains all the operations in the search space $\mathcal{S}$: zero, skip connection, $1\times1$ convolution, $3\times3$ convolution, and $3\times3$ average pooling. 

The random horizontal flipping, random cropping with padding, and normalization are used for data augmentation. We train the supernet with Nesterov momentum SGD for 30 epochs. The weight decay is 0.0005 and the momentum is 0.9. We decay the learning rate from 0.025 to 0.001 with cosine scheduler and clip the gradient by 5. The batch size is set to 64.

\subsection{DARTS-based search space}

The supernet consists of 8 cells, where the third and sixth cells are the reduction cells that connects $N=2$ normal cells in each block. Each cell has 2 inputs(the outputs of the previous cell and the cell before the previous cell) and $V=4$ intermediate nodes connected with $E=14$ edges. 
The search space contains zero, skip connection, $3\times3$ and $5\times5$ separable convolutions, $3\times3$ and $5\times5$ dilated separable convolutions, $3\times3$ max pooling and $3\times3$ average pooling. 

In the searching stage, the supernet is trained on CIFAR-10 with Nesterov momentum SGD for 50 epochs. The other hyper-parameters and the data augmentation methods remain the same as in section A.1. The architecture is shown in Figure \ref{fig:arch}. 

In the evaluation on CIFAR-10, a deeper subnet with $N=3$ is trained for totally 600 epochs with a warmup of 5 epochs. The batch size is 96. The other hyper-parameters remain the same as those used for searching. Moreover, we use additional enhancements including cutout, path dropout of probability 0.2 and auxiliary towers with weight 0.4.

In the evaluation on ImageNet, $N$ changes to 4. Data augmentation include random cropping, color jittering, lighting, random horizontal flipping and normalization. The subnet is trained for 250 epochs with a warmup of 5 epochs, a batch size of 128, a momentum of 0.9 and a weight decay of 0.00003. The learning rate decay from 0.1 to 0 following a cosine scheduler.

\section{Ablation Study}

We explore the influence of removing the selection via IIMs and the shrinking pipeline, denoted as S-DARTS and I-DARTS respectively. The ablation is done on NAS-Bench-201 search space, and the parameters of S-DARTS keeps the same as IS-DARTS. The results is shown in Table \ref{tab:ablation}. 

\begin{table}[h]
  \caption{Ablation Study on the pipelines under NAS-Bench-201.}
  \label{tab:ablation}
  \resizebox{\linewidth}{!}{
  \begin{tabular}{ccccccc}
    \toprule
        \multirow{2}{*}{Method} & \multicolumn{2}{c}{CIFAR-10} & \multicolumn{2}{c}{CIFAR-100} & \multicolumn{2}{c}{ImageNet-6-120} \\ \cline{2-7}
        ~ & valid & test & valid & test & valid & test \\
    \midrule
        IS-DARTS & 91.55 & 94.36 & 73.49 & 73.51 & 46.37 & 46.34 \\ 
        I-DARTS & 90.59 & 93.31 & 70.05 & 69.83 & 43.78 & 43.84 \\ 
        S-DARTS & 68.29 & 70.92 & 38.57 & 38.97 & 18.87 & 18.41 \\ 
    \bottomrule
  \end{tabular}
  }
\end{table}

\begin{figure}[h]
  \centering
  \includegraphics[width=\linewidth]{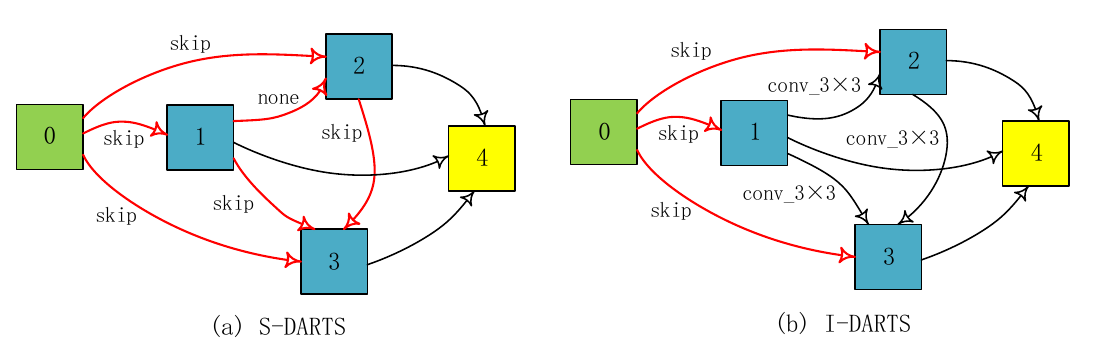}
  \caption{Architecture found by I-DARTS and S-DARTS under NAS-Bench-201 search space.}
  \label{fig:ablation}
\end{figure}

\begin{figure*}[t]
  \centering
  \includegraphics[width=\linewidth]{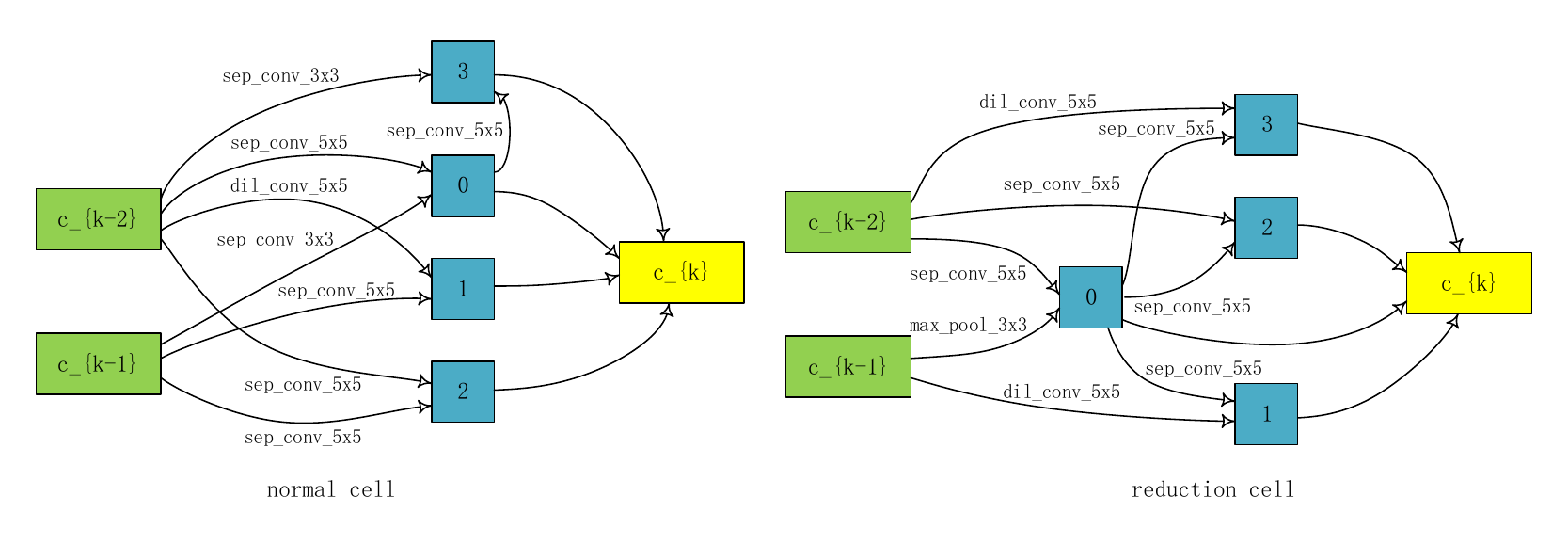}
  \caption{Normal cell and reduction cell found by IS-DARTS on CIFAR-10 under DARTS search space.}
  \label{fig:arch}
\end{figure*}

S-DARTS fails to resolve the fundamental problem of performance collapse with only the shrinking pipeline and ends up with the architecture shown in Figure \ref{fig:ablation}(a), dominated by parameter-free operations. The final architecture of I-DARTS is Figure \ref{fig:ablation}(b), which shows some improvements in the middle of the cells. Wrong decision occurs on all the edges connected to the input of the cells, in that the skip connections which directly bypass all the information extracted by the previous cell is naturally superior to the other operations. In contrast, the shrinking pipeline of IS-DARTS weakens the role of skip connections and makes them comparable with the other operations by steps.

We also conduct ablation studies on different shrink rates under DARTS search space. We adjust the interval epoch to ensure that shrinking only happens in the last 20 epochs. The results are shown in Table \ref{tab:ablation2}. When the shrink rate fluctuates within a reasonable range, the evaluation performance is basically consistent. When the shrink rate is close to 0, the supernets lack sufficient training in less interval epochs. When the shrink rate is close to 1, the gap of supernets in adjacent steps become larger. Both situations worsen the performance. Specially, when the shrink rate is 1, IS-DARTS degrades to I-DARTS.

\begin{table}[h]
  \caption{Ablation Study on different shrink rates under DARTS search space.}
  \label{tab:ablation2}
  \resizebox{\linewidth}{!}{
  \begin{tabular}{c|c|c|c|c|c|c}
	\hline
    shrink rate & 1 (I-DARTS) & 0.5 & 0.25 & 0.143 & 0.1 & 0.05 \\
    interval epoch & / & 18 & 6 & 3 & 2 & 1 \\
    test error & 2.98 & 2.87 & 2.51 & 2.56 & 2.58 & 2.95 \\
    \hline
  \end{tabular}
  }
\end{table}

\section{Extra Visualization}

\begin{figure*}[h]
\centering
\includegraphics[width=\linewidth]{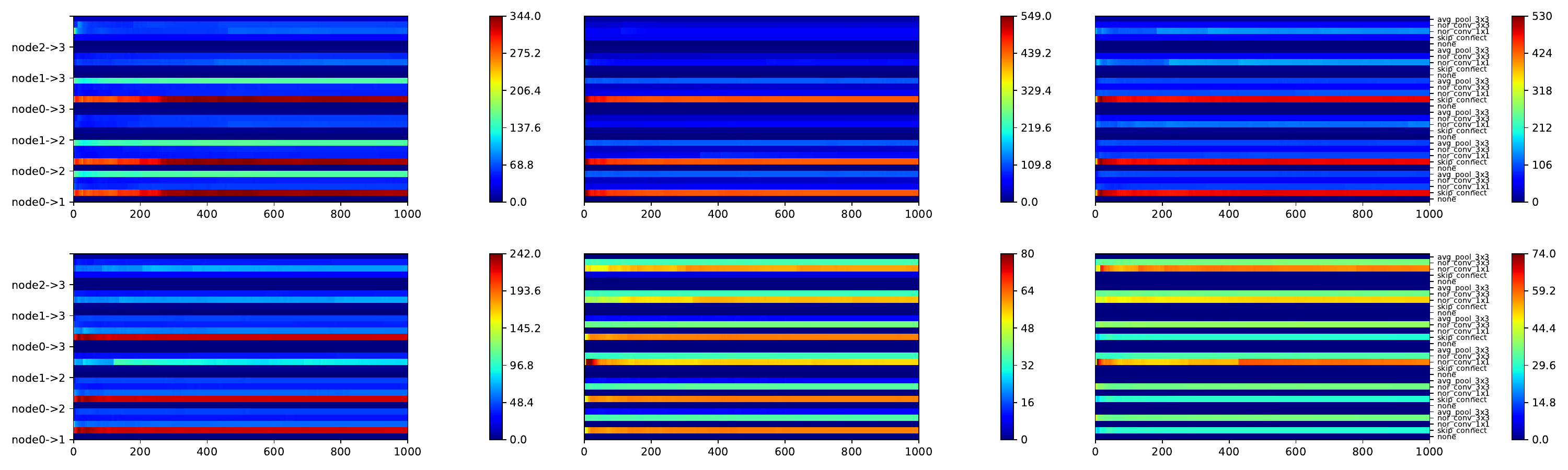}
\caption{The IIMs of operations on all edges at different epochs in the searching stage on NAS-Bench-201 and CIFAR-100.}
\label{fig:exvisual1}
\end{figure*}

We conduct IS-DARTS on NAS-Bench-201 benchmark and CIFAR-100 dataset and achieve 70.71\% validation accuracy and 71.11\% test accuracy. We visualize the IIMs during searching in the same way as Figure \ref{fig:visualization}. Figure \ref{fig:exvisual1} illustrates that IS-DARTS stabilizes with approximately 500 samples.

\begin{figure*}[h]
\centering
\includegraphics[width=\linewidth]{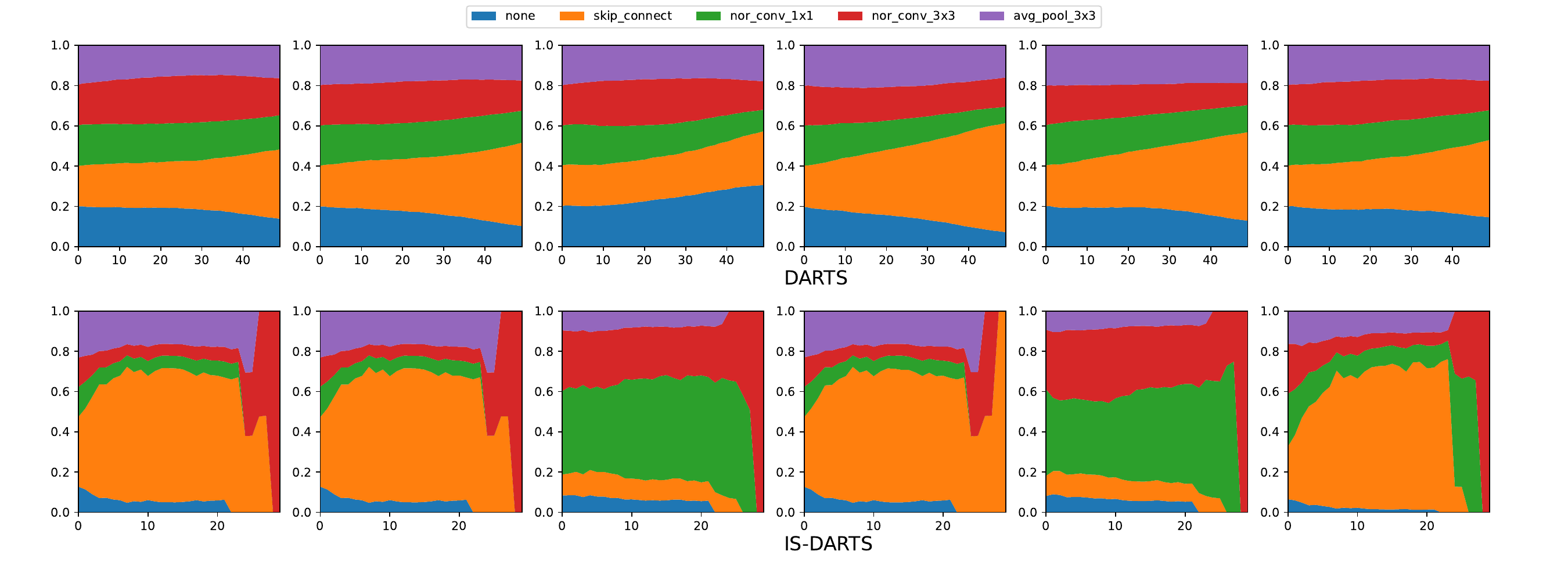}
\caption{The softmax of the architecture parameters in DARTS and the smoothed softmax of information measurements in IS-DARTS under NAS-Bench-201.}
\label{fig:exvisual2}
\end{figure*}

Following Fig. 3 in FairDarts, we visualize the softmax of the architecture parameters in DARTS and the smoothed softmax of information measurements in IS-DARTS in Figure \ref{fig:exvisual2}. It is clearly demonstrated that the IIMs of parameter-free candidates drop sharply once a parameterized candidate is discarded, and performance collapse is resolved in consequence.

\end{document}